\documentclass[nohyperref]{article}

\usepackage{microtype}
\usepackage{graphicx}
\usepackage{subfigure}
\usepackage{booktabs} 
\usepackage{enumitem}

\usepackage[accepted]{icml2022}

\usepackage{amsmath}
\usepackage{amssymb}
\usepackage{mathtools}
\usepackage{amsthm}

\usepackage[capitalize,noabbrev]{cleveref}

\theoremstyle{plain}

\theoremstyle{definition}

\theoremstyle{remark}

\usepackage[textsize=tiny]{todonotes}

\icmltitlerunning{Advantages of Dense-Vector to One-Hot Encoding in Out-of-Scope Detection}

\begin{document}

\twocolumn[


\icmltitle{Exploring the Advantages of Dense-Vector to One-Hot Encoding \linebreak of Intent Classes in Out-of-Scope Detection Tasks}



\icmlsetsymbol{equal}{*}

\begin{icmlauthorlist}
\icmlauthor{Claudio Pinhanez}{equal,yyy}
\icmlauthor{Paulo Cavalin}{equal,yyy}

\end{icmlauthorlist}

\icmlaffiliation{yyy}{IBM Research, Brazil}

\icmlcorrespondingauthor{Claudio Pinhanez}{claudio@pinhanez.com}


\vskip 0.3in
]



\printAffiliationsAndNotice{}  

\begin{abstract}
This work explores the intrinsic limitations of the popular one-hot encoding method in classification of intents when detection of out-of-scope (OOS) inputs is required. Although recent work has shown that there can be significant improvements in OOS detection when the intent classes are represented as dense-vectors based on domain specific knowledge, we argue in this paper that such gains are more likely due to advantages of dense-vector to one-hot encoding methods in representing the complexity of the OOS space. We start by showing how dense-vector encodings can create OOS spaces with much richer topologies than one-hot encoding methods. We then demonstrate empirically, using four standard intent classification datasets, that knowledge-free, randomly generated dense-vector encodings of intent classes can yield massive, over 20\% gains over one-hot encodings, and also outperform the previous, domain knowledge-based, SOTA of one of the datasets. We finish by describing a novel algorithm to search for good dense-vector encodings and present initial but promising experimental results of its use.

\end{abstract}

\section{Introduction}
Dense representations of inputs to machine learning (ML) models, often referred to as \textit{input embeddings}, have been one of the key drivers of the massive improvements of performance of most NLP applications in the last 10 years. However, the use of similar \textit{dense-vector} representations for the output classes, or \textit{output embeddings}~\cite{yu2010attribute,rohrbach2011evaluating,kankuekul2012online,akata2016}, is quite uncommon, except in scenarios of \textit{zero-shot learning}~\cite{pmlr-v37-romera-paredes15} where embeddings are used to encode non-observed latent classes. 

The most used method to represent $c>0$ different intent classes is still to encode them as \textit{one-hot} vectors , i.e. $c$-dimensional vectors which are all filled with 0s (zeros) except in the position corresponding to the class of the intent label, which is filled with. Therefore classes are represented as equally-distant points in a $c$-dimensional space and classification is performed by computing the distance to the closest one-hot vector, often using the $softmax$ function.

Recent works~\cite{cavalin2020improving,pinhanez-etal-2021-using} have shown that dense-vector representations of intent classes based on domain knowledge can improve intent classification accuracy. In particular, such representations have shown to produce impressive gains in the detection of the inputs to the machine learning system which are outside of the scope of the intent classes, often referred to as \textit{out-of-scope (OOS)} or \textit{out-of-domain (OOD)} samples~\cite{LeeNEURIPS2018,Vyas_2018_ECCV, LeeICLR2018, chen2020robust}. Notice that a classification task without OOS detection in fact assumes a \textit{closed world} scenario, while the full version, which detects both the correct class for \textit{in-scope (IS)} samples or whether they are OOS samples, corresponds to the \textit{open world} case. Most real applications are in open world contexts, including almost any practical use of intent classification. Here we consider only the case where no examples of the OOS class are provided for training, although some are available for testing.

The main argument of this paper is that such large accuracy gains in OOS~detection are more likely due to advantages of dense-vector to one-hot encoding of intent classes than to the use of domain knowledge. We start by showing that, in OOS detection, the complexity of the spaces representable by one-hot encodings is quite limited when compared to dense-vector ones. This argument is made based on the number of different topological spaces enabled by each type of encoding. We show that one-hot encodings of $c$ classes can create only $c$ types of topologically-distinct spaces, while the number of different spaces enabled by dense-vector encodings is larger than $c^2$.

We follow by presenting the results of some experiments in four intent classification public datasets, where dense-vectors yielded 20\% to 40\% improvements in the accuracy of OOS detection. We did this by randomly generating dense-vector encodings. Our experiments also achieved better results than a previous SOTA method based on domain knowledge which encode intent classes using \textit{word graphs}~\cite{cavalin2020improving}. We did not obtain similar improvements in IS intent classification, where the one-hot \textit{softmax}-based technique was as good as any dense-vector encoding.


However, those encodings were found through random trial-and-error, a highly inefficient process. We close the paper proposing an algorithm which searches the space of encodings, looking for an optimal combination of a set of dense-vectors and a classification neural network. In initial experiments, the algorithm improved by 10\% the OOS detection rate of the best random dense-vector encoding used in the previous experiment, and found a representation as good as one-hot \textit{softmax} for intent classification.

Given those results, we propose that intent classification with and without OOS detection should be regarded as quite different problems with distinct methods to solve them. This work makes the case that the use of more powerful representation systems, i.e., dense-vectors, may have formidable performance impacts in the classification with OOS detection case, and that a possible explanation for the success is their ability to represent more complex topological spaces.

\section{Related Work}

This paper focuses on the task known as \emph{out-of-domain sample detection}~\cite{Tan2019} or \emph{out-of-distribution sample detection}~\cite{LeeNEURIPS2018,Vyas_2018_ECCV, LeeICLR2018, chen2020robust}; a good survey can be found in~\cite{yang2021generalized}. Many of the existing approaches for OOS detection rely on adapting the training algorithm by changing the loss function of a neural network~\cite{LeeICLR2018}; by generating ensembles of classifiers~\cite{Vyas_2018_ECCV,ShaledNEURIPS2018}; by exposing the model to dversarial, crafted inlier and outlier examples~\cite{chen2020robust, li-etal-2021-kfolden}; or by including additional OOS examples either in an unsupervised way~\cite{Yu_2019_ICCV,Tan2019} or by generating them~\cite{Vernekar2019}. Another approach is to transform and improve the \textit{softmax} outputs~\cite{hendrycks17baseline,LiangICLR2018,techapanurak2019cosine}; or to generate an additional classifier to measure the confidence of the ML classifier, following an assumption that processing OOS samples have low-confidence scores~\cite{Ryu2017,Ryu2018,devries2018learning, LeeNEURIPS2018}.

Instead, our approach focuses on changing the representation of the output layer to match it better with the characteristics of the space of intent classes. In many ways, we explore in the output layer one of the most important advances in machine learning, which is the use of \textit{input embeddings}~\cite{turian2010word,mikolov2013efficient,pennington2014glove,dossantos2014deep}. In particular, we look into a new use for \textit{output or class embeddings}, which have been explored before in other contexts. In particular, in~\emph{zero-shot learning}~\cite{pmlr-v37-romera-paredes15}, class embeddings have been used as a tool that makes it possible building a solution for the problem. Zero-shot learning is based on the identification and addition of new classes to a classifier with no reliance in input samples. With class embeddings, new classes can be added to the system by simply generating an embedding with the proper configuration of an unseen class, to encapsulate the knowledge of the new concepts~\cite{palatucci2009,socher2013,Akata_2015_CVPR,akata2016}. Zero-shot learning is a problem closely related to OOS detection but they differ on the criterion of success: in the former, the accuracy of assigning inputs to the previously unknown classes; in the latter, the accuracy of identifying inputs which do not belong to any of the known classes. We focus here solely in the OOS detection problem but some of our arguments may also be valid for zero-shot learning.

Recent research has focused in using class embeddings to enhance a ML classifier by encapsulating additional high-level knowledge related to the classes, such as in~\cite{cavalin2020improving, pinhanez-etal-2021-using}. 
In~\cite{cavalin2020improving} the classes were represented by a keyword extracted from the class training examples followed by the embedding of the corresponding \textit{word graph}. However, word graphs tend to repeat the class examples  with a different structure, thus are far from ideal to produce proper class embeddings.  In~\cite{pinhanez-etal-2021-using} the hierarchical taxonomy of the classes, as understood by the system developers, was mined from the documentation of the system and used to create class embeddings. Although the latter approach seems promising, such taxonomies might not be available in many cases, what limit its applicability. Notice that those two approaches excelled particularly in OOS detection.


This work explores further the use of class embeddings by looking into the properties of dense-vectors themselves, independent of the presence of knowledge. Our key baseline for comparison is the traditional one-hot encoding methods. Some previous works have explored the difference between one-hot and dense-vectors, such as~\cite{rodriguez2018beyond}, which found higher rates of convergence for the latter. Output embeddings have also been explored in the context of \textit{multi-class classification} problems~\cite{amit2007uncovering,weinberger2009large,weston2010large,akata2015label} and \textit{large-scale recognition}~\cite{srivastava2013discriminative,deng2014large,xiao2014error,yan2015hd,lin2015deep}. We are not aware of works focusing on OOS detection using knowledge-free class embeddings as described in this paper.

\section{Class Encoding  for OOS Detection}
\label{sec:class_encoding_oos}

Following the notation of~\cite{cavalin2020improving}, an \emph{intent classification} method is a function $D$ which maps a set of sentences (potentially infinite) $S=\{s_{1},s_{2},...\}$ into a finite set of classes $\Omega=\{\omega_{1},\omega_{2},...,\omega_{c}\}$:
\begin{equation}
    D:S\rightarrow\Omega \hspace{5mm} D(s)=\omega_{i} 
\end{equation}

In many practical situations of intent classification, it is also necessary to determine whether a sentence $s$ does not belong to any of the classes, what is often referred to as \textit{out-of-scope (OOS) detection} or \textit{out-of-domain (OOD) detection}. This can be represented by expanding $\Omega$ to the set $\bar{\Omega} = \Omega \cup \{o\}$ where $o$ represent the class OOS of samples. Following, an \textit{intent classification with OOS detection} method $\bar{D}:S\rightarrow \bar{\Omega}$ is defined by:
\begin{equation}
   \bar{D}(s) =  \left\{ 
   \begin{array}{ll} 
    D(s)=\omega_{i} &\mbox{~if~in-scope} \\ 
    o &  \mbox{~if~out-of-scope}
    \end{array}
    \right.
\end{equation}

An input embedding $\xi:S\rightarrow\mathbb{R}^{n}$ is often used, mapping the space of sentences $S$ into a vector space  $\mathbb{R}^{n}$, and defining a classification function $\bar{E}:\mathbb{R}^{n}\rightarrow\bar{\Omega}$ such as $\bar{D}(s)=\bar{E}(\xi(s))$. 
In typical intent classifiers, $\bar{E}$ is usually composed of a function $M$ which computes the $z=(z_1, z_2, ..., z_c)$  likelihood of $s$ being in each class $\omega_i$, typically in a finite range such as $[-1,1]$, followed by a \textit{class encoding} function $\bar{C}$ which maps the likelihood results into the classes in $\bar{\Omega}$.
\begin{equation}
S \overset{\xi}{\rightarrow} \mathbb{R}^{n} \overset{M}{\rightarrow} [-1,1]^{c} \overset{\bar{C}}{\rightarrow} \bar{\Omega}
\end{equation}

A common way to implement $\bar{C}$, denoted here as $\bar{C}_{max}$, is to verify whether any coordinate $z_i$ is greater than a threshold $0 \leq \theta < 1$, and, if so, to map it into the $\omega_i$ associated with the maximum $z_i$ value; otherwise, $\bar{C}_{max}$ maps it into $o$. 
\begin{equation}
   \bar{C}_{max}(z,\theta) =  \left\{ 
   \begin{array}{ll} 
    \mbox{argmax~} z_i &\mbox{~if~} \max z_i > \theta  \\ 
    o      &  \mbox{~otherwise}
    \end{array}
    \right.
\end{equation}

Probably the most common method for class encoding is to use the $softmax$ function where the likelihood components are normalized with the exponential funcion, denoted here as $\bar{C}_{softmax}$. 
\begin{equation}
\sigma(z)_i = \frac{e^{z_i}}{\sum_{j=1}^{c}e^{z_j}} \mbox{~~~thus~~~} \sum_{i=1}^{c} \sigma(z)_i =1
\end{equation}
\begin{equation}
\bar{C}_{softmax}(z,\theta) =  \left\{ 
   \begin{array}{ll} 
   \mbox{argmax~} \sigma(z)_i &\mbox{~if~} \max \sigma(z)_i > \theta  \\ 
    o      &  \mbox{~otherwise}
    \end{array}
    \right.
\end{equation}



Notice that $\bar{C}_{max}$ and $\bar{C}_{softmax}$ can be seen as computing the $min$-based distance $d_{min}$ of $z=(z_1, z_2, ... z_c)$ to \textit{one-hot} vectors $h_i=(0,0,....,1,...,0,0)$ where the $1$ value is in the i-th position of $h_i$. In this paper we compare those methods with approaches based on Euclidean distance.  If instead of $min$ we use the \textit{Euclidean distance} $d$ to the one-hot vectors $h_i$, we obtain a function which we call  $\bar{C}_d$:
\begin{equation}
   \bar{C}_{d}(z,\theta) =  \left\{ 
   \begin{array}{ll} 
    \mbox{argmin~} d (z_i,h_i) &\mbox{if~} \min  d(z_i,h_i) \leq \theta  \\ 
    o      &  \mbox{otherwise}
    \end{array}
    \right.
\end{equation}

Since all previous methods consider which intent class is closer, for a particular distance, to the one-hot vectors, we refer to them as \textit{one-hot class encoding} methods. An alternative approach is to consider each class $\omega_i$ as represented by the points closer to a given \textit{dense-vector} $r_i=(r_{i1}, r_{i2}, ... r_{ic}) \in [-1,1]^c$. We call this function $\bar{C}_r$, defining a typical \textit{dense-vector encoding} method.
\begin{equation}
   \bar{C}_{r}(z,\theta) =  \left\{ 
   \begin{array}{ll} 
    \mbox{argmin~} d (z_i,r_i) &\mbox{if~} \min d(z_i,r_i) \leq \theta  \\ 
    o      &  \mbox{otherwise}
    \end{array}
    \right.
\end{equation}

The use of dense-vector encoding opens up the exploration of different dimensions beyond $c$-dimensional encodings as the output of the likelihood function $M$. In fact, any dimension $p > 0$ can be used, and in this case each class $\omega_i$ is represented by the points closer to $r_i^p=(r_{i1}^p, r_{i2}^p, ... r_{ic}^p) \in [-1,1]^p$, defining the function $\bar{C}_r^p$.
\begin{equation}
   \bar{C}_{r}^{p}(z^p,\theta) =  \left\{ 
   \begin{array}{ll} 
    \mbox{argmin~} d (z_i^p,r_i^p) &\mbox{if~} \min d(z_i^p,r_i^p) \leq \theta  \\ 
    o      &  \mbox{otherwise}
    \end{array}
    \right.
\end{equation}

Dense-vector encoding methods can use a variety of ways to generate the $r_i$ points to represent the intent classes. We explore experimentally in this paper both \text{random} methods to generate such points and the knowledge-informed method based on word graphs described in~\cite{cavalin2020improving}. But to better understand the differences between using one-hot and dense-vector encodings, we first discuss the different representational topological expressivity of each class encoding method to handle different cases of component connectiveness of the OOS space.

\section{Class Encoding Topologies}
\label{sec:class_encodings_topologies}

In this section we investigate the differences of the class encoding methods by considering first a simplified 2D scenario where the goal is to determine whether a sentence $s$ belongs to one of two classes, $\omega_1$ or $\omega_2$, or is out-of-scope ($o$). The generalization to high-dimensional spaces uses similar arguments and is detailed in Appendix~\ref{app:class_encoding_topologies_higher_dimensions}.

\begin{figure}[t!]
    \centering
    \includegraphics[width=8cm]{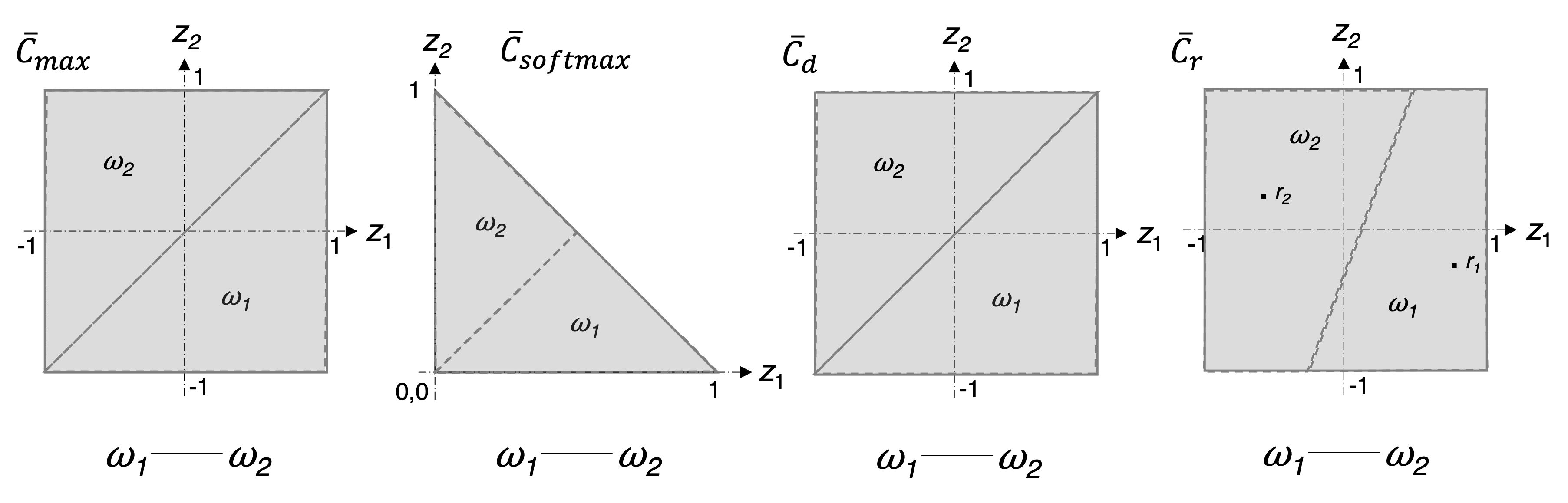}
    \caption{The same topology is generated by $\bar{C}_{max}$, $\bar{C}_{softmax}$, $\bar{C}_{d}$, and $\bar{C}_r$ in 2D when there is no OOS detection ($\theta=0$).}
    \label{fig:topologies_without_oos}
\end{figure}

To simplify the analysis we do not consider here special, limit cases where the intersection of two components degenerates to a single point or to a tangent line. Considering this, let us start by examining the simplest situations where there is no OOS detection. This corresponds to the case where $\theta=0$ in the previous formulas. Figure~\ref{fig:topologies_without_oos} shows that, in those cases, all the different functions $\bar{C}$ described before can only generate the same topology in which the $\omega_1$ and the $\omega_2$ components are connected. 

\begin{figure}[t!]
    \centering
    \includegraphics[width=8cm]{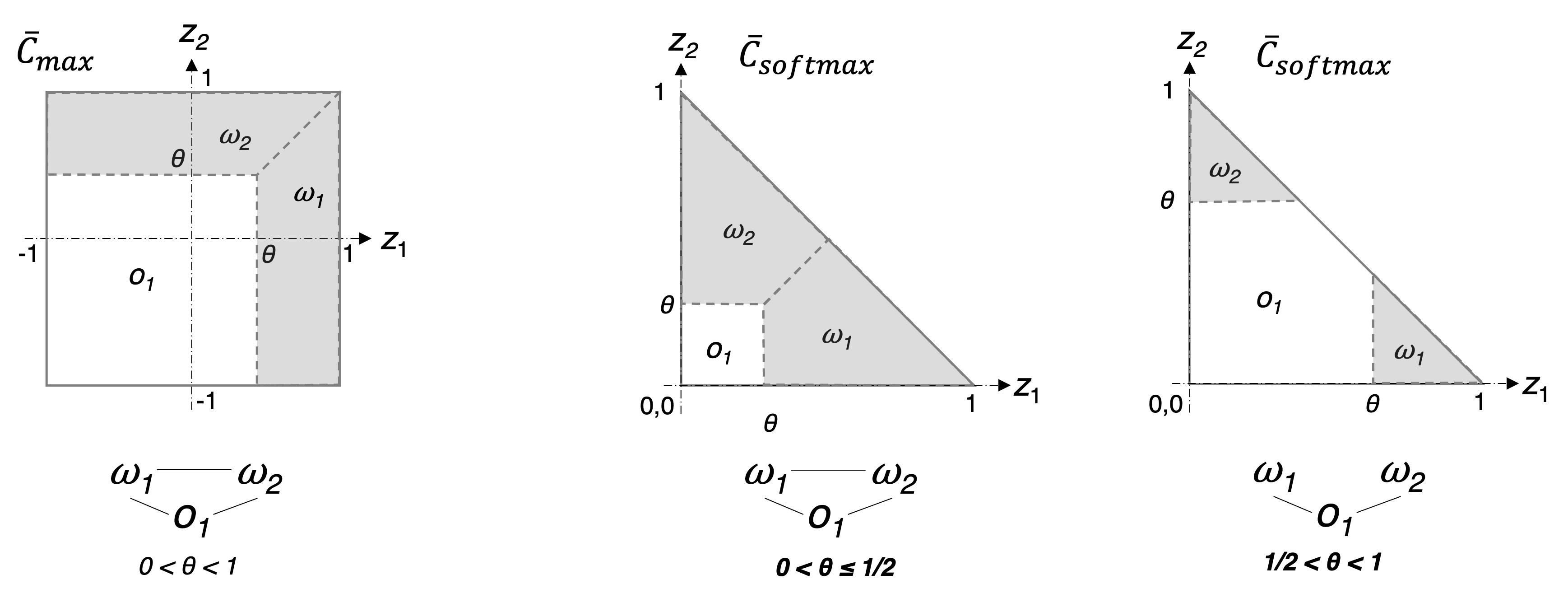}
    \caption{The only $\bar{C}_{max}$ topology  and the two topologies of $\bar{C}_{softmax}$ in 2D.}
    \label{fig:one-hot_max_normal_topologies}
\end{figure}

However, when there is OOS detection, that is, $\theta>0$, a different topological landscape emerges. The leftmost 2D space of fig.~\ref{fig:one-hot_max_normal_topologies} illustrates the case of the $\bar{C}_{max}$ function where the class $\omega_1$ maps into a trapezium positioned between $\theta \leq z_1 \leq 1$, bound by the main diagonal of the first quadrant as shown. The class $\omega_2$ similarly maps into a reflected trapezium and the OOS class $o_1$ occupies a square defined by $-1 \leq z_1 < \theta$ and $-1 \leq z_2 < \theta$. We represent schematically the topology of the connected components defined by $\bar{C}_{max}$ as the leftmost bottom diagram of fig.~\ref{fig:one-hot_max_normal_topologies}. It shows that the three components are pair-wise connected, that is, share a non-trivial border in this space.

Although quite similar to $\bar{C}_{max}$, the $\bar{C}_{softmax}$ function can represent two distinct topologies, as shown in the central and rightmost parts of fig.~\ref{fig:one-hot_max_normal_topologies}. This is an effect of the normalization process which maps all points into a triangle in the first quadrant. The first topology is identical to the case of $\bar{C}_{max}$, and it happens if $0 < \theta \leq 1/2$. However, if $\theta > 1/2$, the square of the OOS component $o_1$ divides the two intent classes into two non-connected triangles, yielding a new topological configuration where the $\omega_1$ and $\omega_2$ are not connected as shown in~fig.~\ref{fig:one-hot_max_normal_topologies}.

\begin{figure}[t!]
    \centering
    \includegraphics[width=8cm]{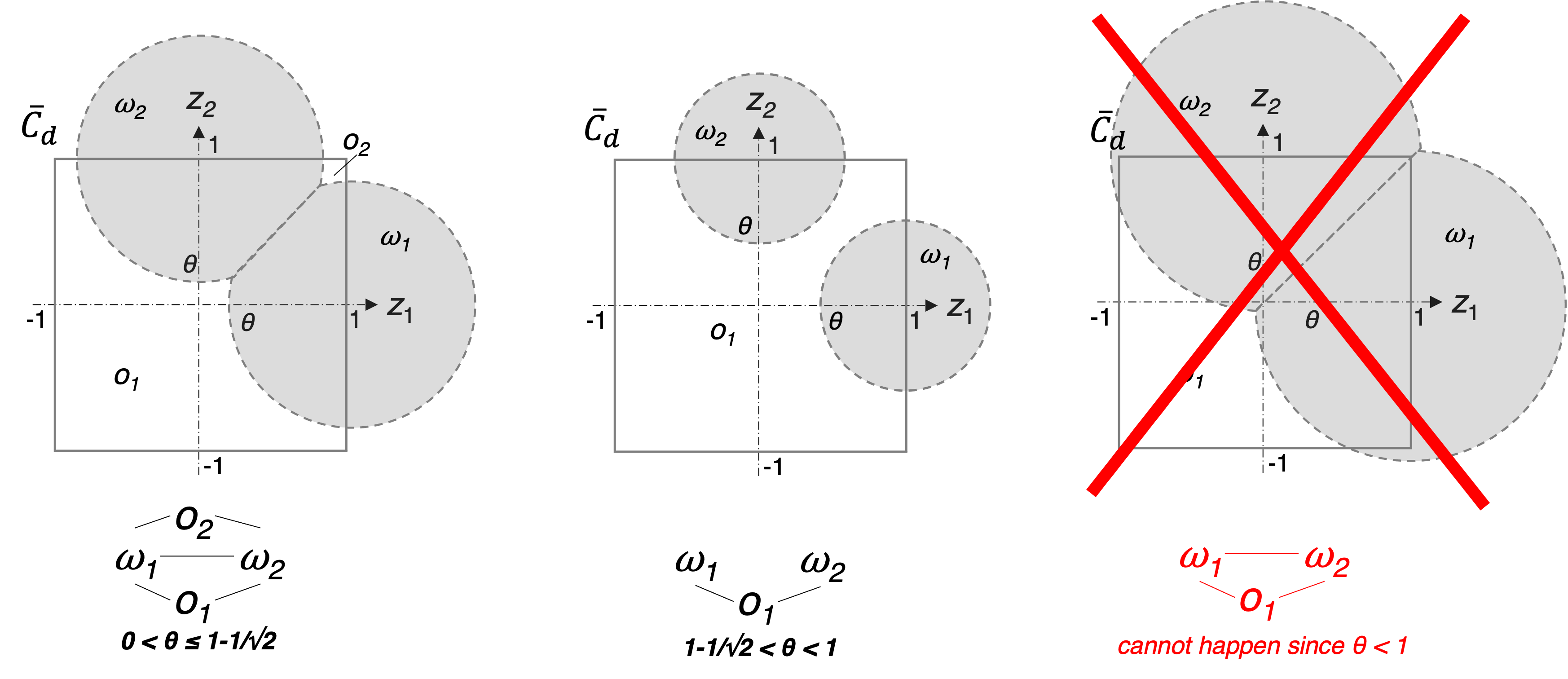}
    \caption{The topologies of $\bar{C}_{d}$ in 2D. The rightmost topology is not allowed because $\theta \leq 1$.}
    \label{fig:one-hot_distance_topologies}
\end{figure}

Continuing the exploration of one-hot class encodings, fig.~\ref{fig:one-hot_distance_topologies} shows the effects of substituting $max$ with the Euclidian distance $d$, which basically maps $w_1$ and $w_2$ into circles of $1-\theta$ radius centered on the one-hot vectors $(1,0)$ and $(0,1)$. If $ 1-1\sqrt{2} < \theta < 1$ we obtain the case depicted on the center of fig.~\ref{fig:one-hot_distance_topologies}, a topology similar to the second case of $\bar{C}_{softmax}$ where the two intent classes are disconnected and connected to a common OOS component $o_1$. However, when $0< \theta \leq 1-1/\sqrt{2}$ a new topological configuration is enabled, created by a new unconnected OOS component $o_2$ corresponding to points with high values of both $z_1$ and $z_2$, as seen in the leftmost part of fig.~\ref{fig:one-hot_distance_topologies}. Also, as seen in the rightmost part of fig.~\ref{fig:one-hot_distance_topologies}, it is not possible to have this case a topological configuration where the two classes are pair-wise connected but the OOS space is not split into two components. That would require that the threshold $\theta$ to be greater than $1$ what is not possible. Notice that although the \textit{sofmax} and Euclidean distance $d$ in one-hot encoding representations allow for different topologies of OOS spaces, both afford just two different topologies which depend on the value of $\theta$.

\begin{figure}[t!]
    \centering
    \includegraphics[width=8cm]{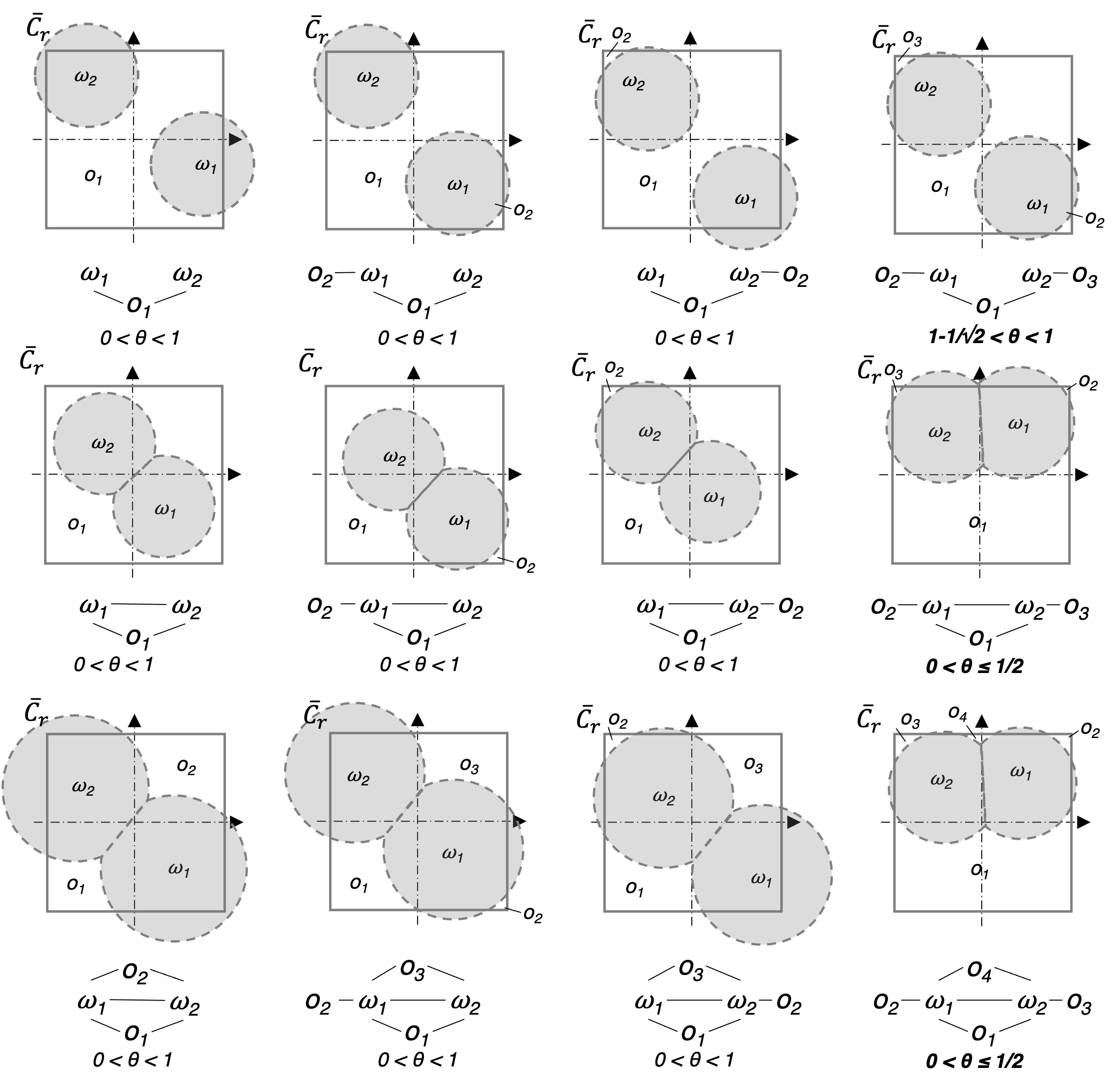}
    \caption{The twelve topologies of  $\bar{C}_{r}$ in 2D.}
    \label{fig:random_distance_topologies}
\end{figure}

Figure~\ref{fig:random_distance_topologies} shows how the use of dense-vector encoding enables many more and richer representations of the output space than the one-hot methods. As depicted, dense-vectors in 2D allow 12~different topological configurations, and the split of the OSS space into 2, 3, and even 4 disconnected components. Also, dense-vector encoding allows for situations where one OOS component is connected to just one of the intent classes, such as in the 8~cases of the two central columns of fig.~\ref{fig:random_distance_topologies}.

This analysis of the 2D scenario shows how potentially limiting is the use of one-hot encoding methods, especially in their ability to represent more complex topological configurations of the OOS space. 
More important, such analysis holds for higher dimensions of $c$ as discussed in detail in~Appendix~\ref{app:class_encoding_topologies_higher_dimensions}.
For the main argument of this paper, it suffices to say that, for any number of classes $c \geq 2$, the one-hot encoding function $\bar{C}_{max}$ defines only one topology, and $\bar{C}_{softmax}$ and $\bar{C}_d$ define exactly $c$ different topologies. However, for dense-vector encoding methods such as $\bar{C}_r$, the number of different topologies increases at least quadratically with $c$, as demonstrated in Appendix~\ref{app:class_encoding_topologies_higher_dimensions}.


\section{Empirical Evaluations of the Encodings}

In the previous section we established that dense-vector encoding methods can represent much more complex OOS components and output spaces than one-hot encoding. We now present empirical evidence that dense-vector methods can outperform significantly one-hot encoding methods in OOS detection tasks and that in tasks without OOS detection the gains are small or non-existent.

\subsection{The Algorithms}
We used only the \emph{Universal Sentence Embeddings (USE)}~\cite{cer-etal-2018-universal} as the main classifier in our experiments.
We consider the baseline in our experiments the traditional classification methods
which represent a given class symbol in the one-hot encoding format. 
For simplicity, in our experiments we considered only the \emph{softmax} function $\bar{C}_{softmax}$ for the final step of the classification. We refer to this algorithm as \textbf{1-hot softmax}.

Next, we used the \textbf{1-hot distance} method which implements one-hot encodings using Euclidean distance, corresponding to the function $\bar{C}_d$. 
The main reason for this is to evaluate the impact of switching from $softmax$ to Euclidean distance, which, as we saw, creates more topologically complex OOS spaces though not increasing the topological count. 


As for dense-vector encodings, we evaluated both knowledge-based and random methods. For domain knowledge-based dense-vector encodings, we used the algorithm described by~\cite{cavalin2020improving}, called here \textbf{word graph}. Basically, the graph embedding algorithm \emph{DeepWalk}~\cite{perozzi2014deepwalk} was used to generate the dense-vector class embeddings based on a graph composed of nodes to represent the classes linked to keywords extracted from the examples in the training set. The resulting class encodings were the graph embeddings associated to the class nodes. 



For the random sampling dense-vector encodings, we employed algorithms identical to the one used in \textit{word graph} except that they used $N$-dimensional random dense-vector class embeddings instead of the word graph embeddings. Classes were represented by $N$-dimensional vectors filled with values which were randomly sampled from an uniform distribution. We refer to the corresponding algorithms as \textbf{R(N)}. For evaluation purposes, we computed a set of random samples for each \textit{R(N)} and considered both the average and minimum value of the metrics across the set of samples.


\subsection{Evaluation Metrics} 
To evaluate the different methods, we used an uniform way to select the key $\theta$ value. We considered the fairest way to do so is to use the \emph{equal error rate (EER)} evaluation metrics employed in~\cite{Ryu2017,Ryu2018,Tan2019,cavalin2020improving,pinhanez-etal-2021-using}. 
In this evaluation metrics, the threshold $\theta$ is set based on the value where the curves of \emph{false acceptance rate (FAR)} and \emph{false rejection rate (FRR)} intersect. 
The former metric corresponds to number of accepted OOS samples divided by the total of OOS samples; the latter represents the ratio between the number of wrongly rejected \textit{in-scope (IS)} cases and the total of IS samples.
Additionally, we also took into account \emph{in-scope error rate (ISER)}, which corresponds to the error considering only IS samples and no rejection, i.e. $\theta = 0$, similar to the class error rate in~\cite{Tan2019}.

\subsection{Experiments and Results}
\label{sec:experiments_results}
The different algorithms were evaluated in four distinct intent classification datasets, all of them containing tagged OOS examples.
These datasets comprise the three \textit{HINT3} intent recognition problems~\cite{arora-etal-2020-hint3}, consisting of challenging small and unbalanced datasets; and the balanced dataset known as \emph{CLINC150}~\cite{larson-etal-2019-evaluation}, referred here as \textit{Larson}, which is considerably larger both in the number of classes and examples. 

\begin{figure*}[t]
    \centering
    \includegraphics[width=5.5cm]{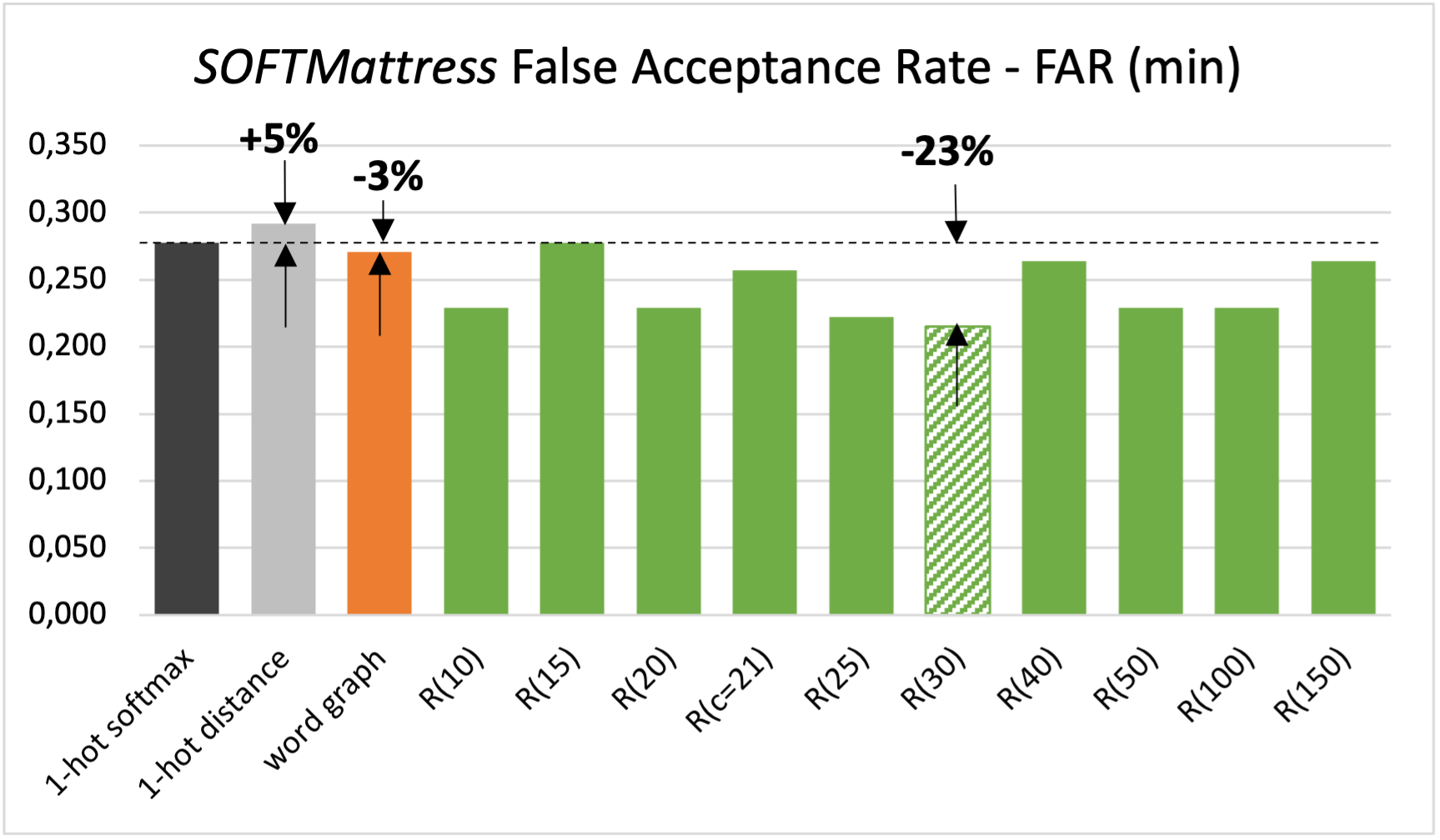}
    \includegraphics[width=5.5cm]{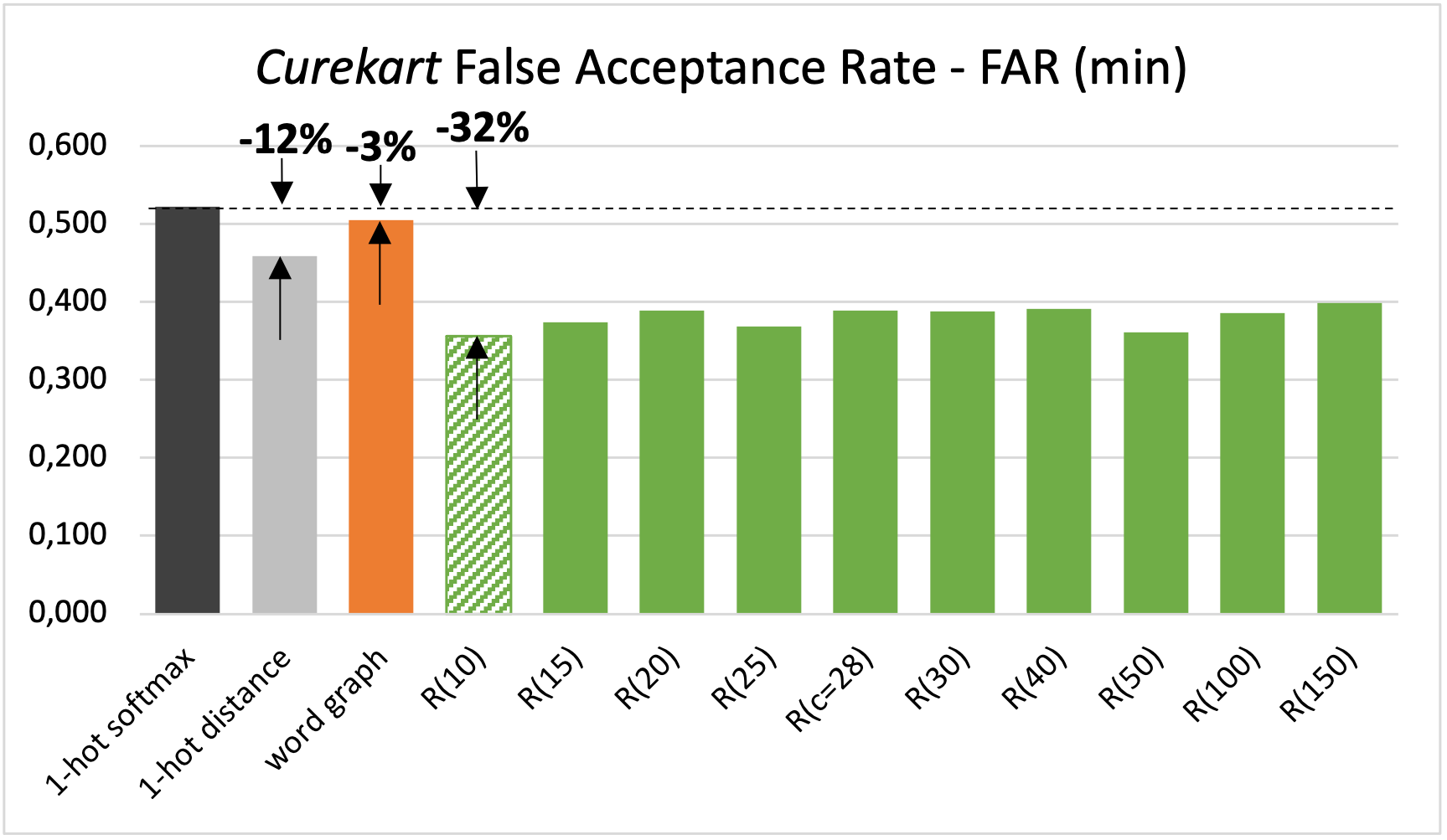} 
    \includegraphics[width=5.5cm]{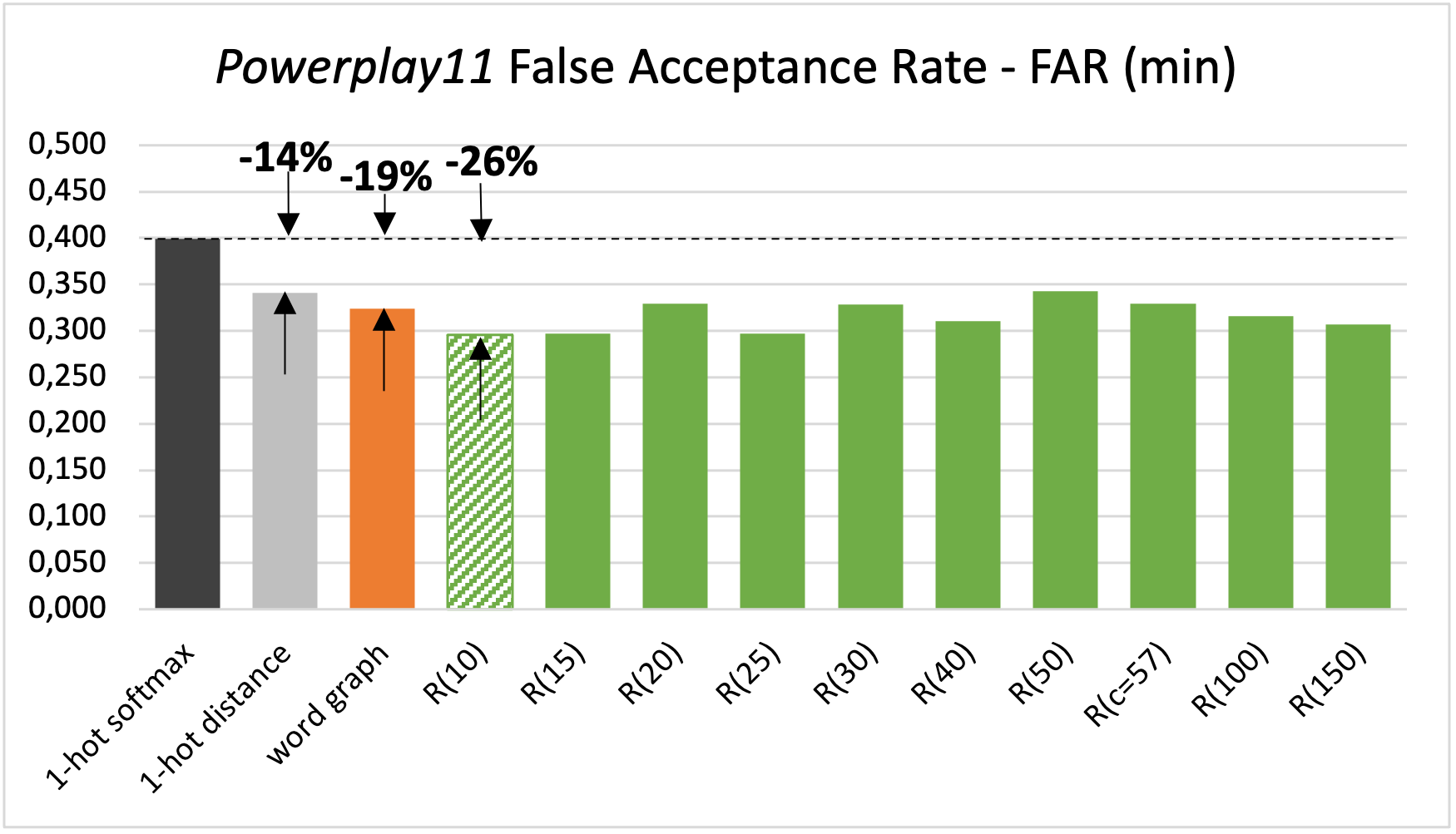}
    \\
    \includegraphics[width=5.5cm]{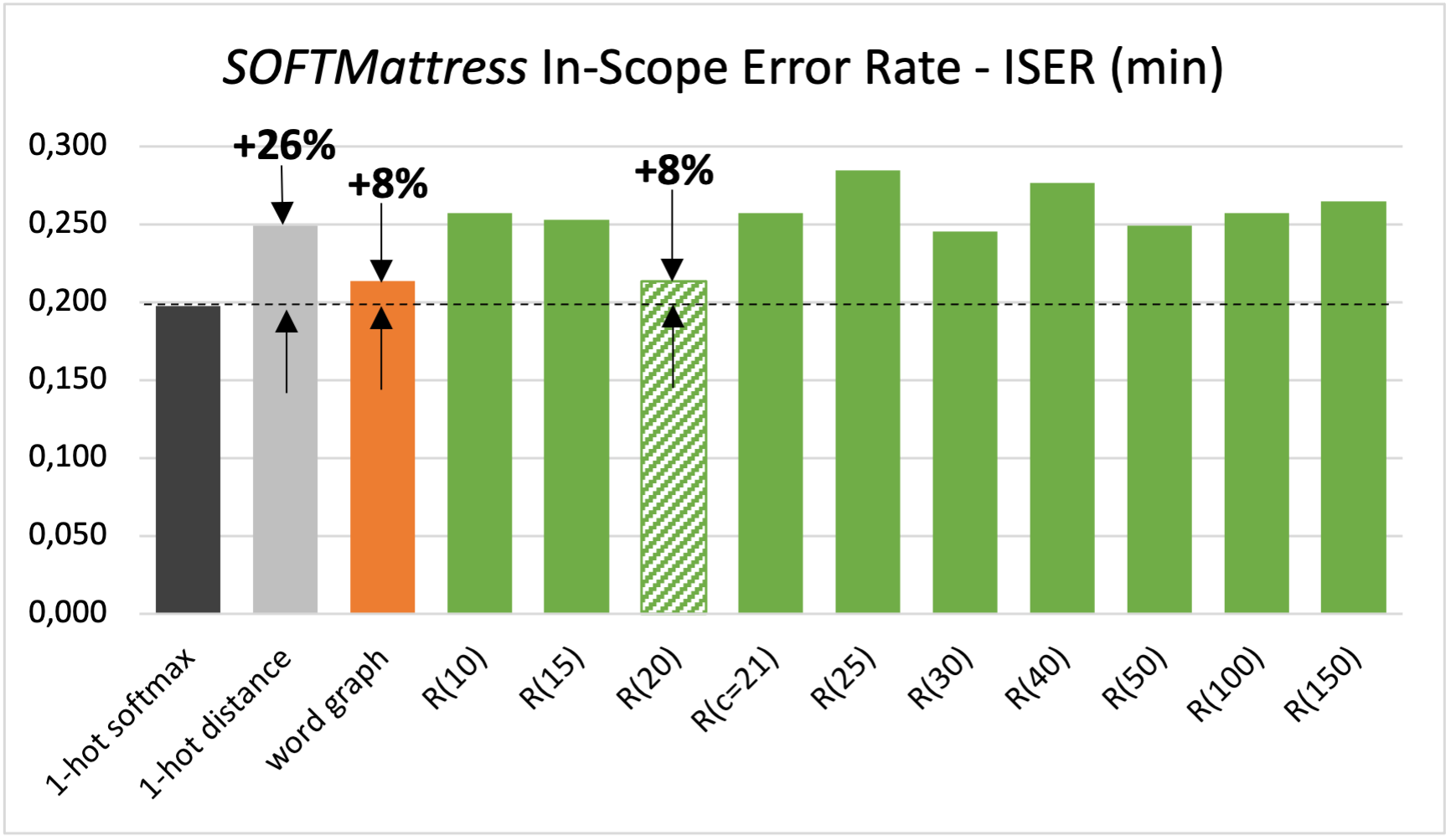} 
    \includegraphics[width=5.5cm]{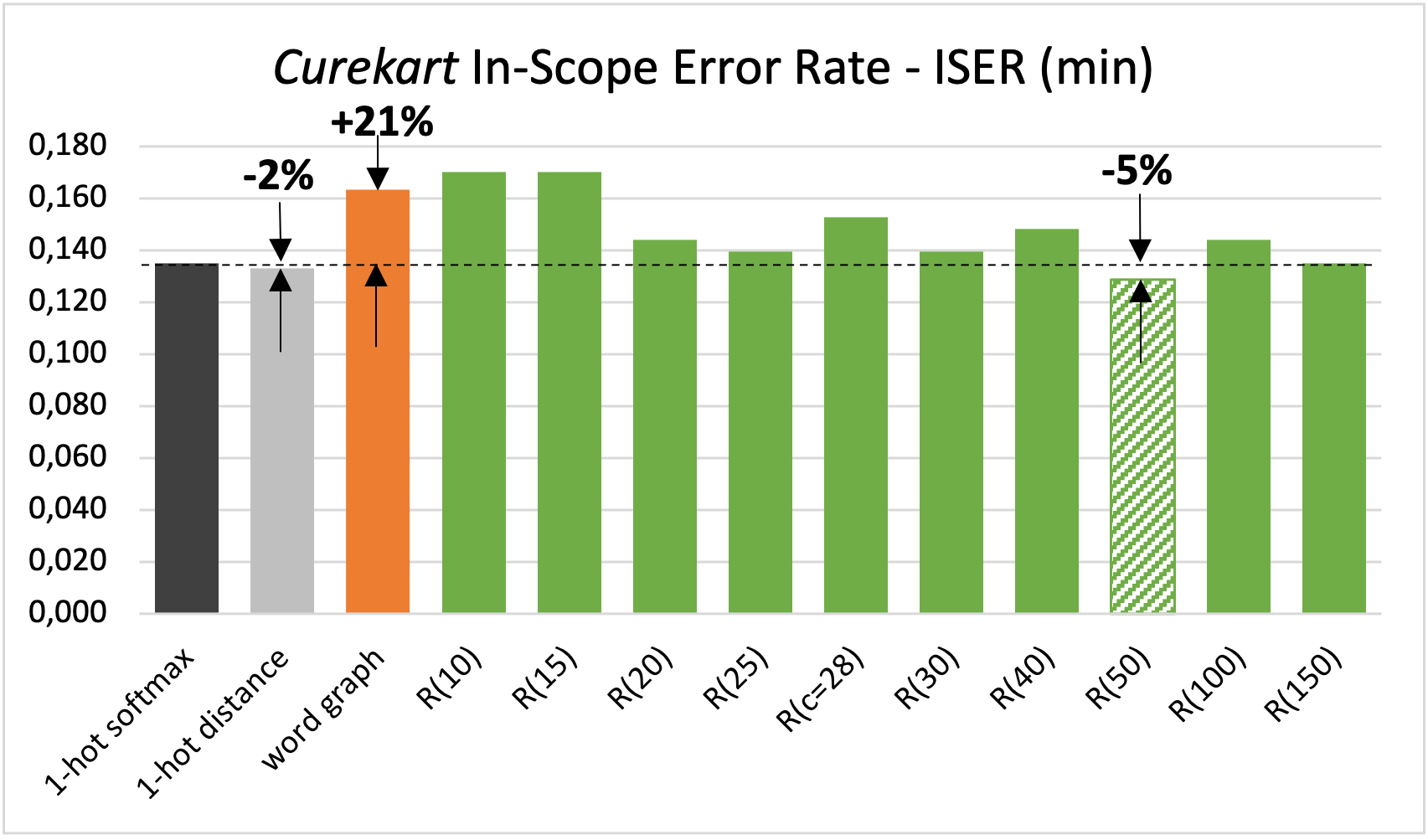}
    \includegraphics[width=5.5cm]{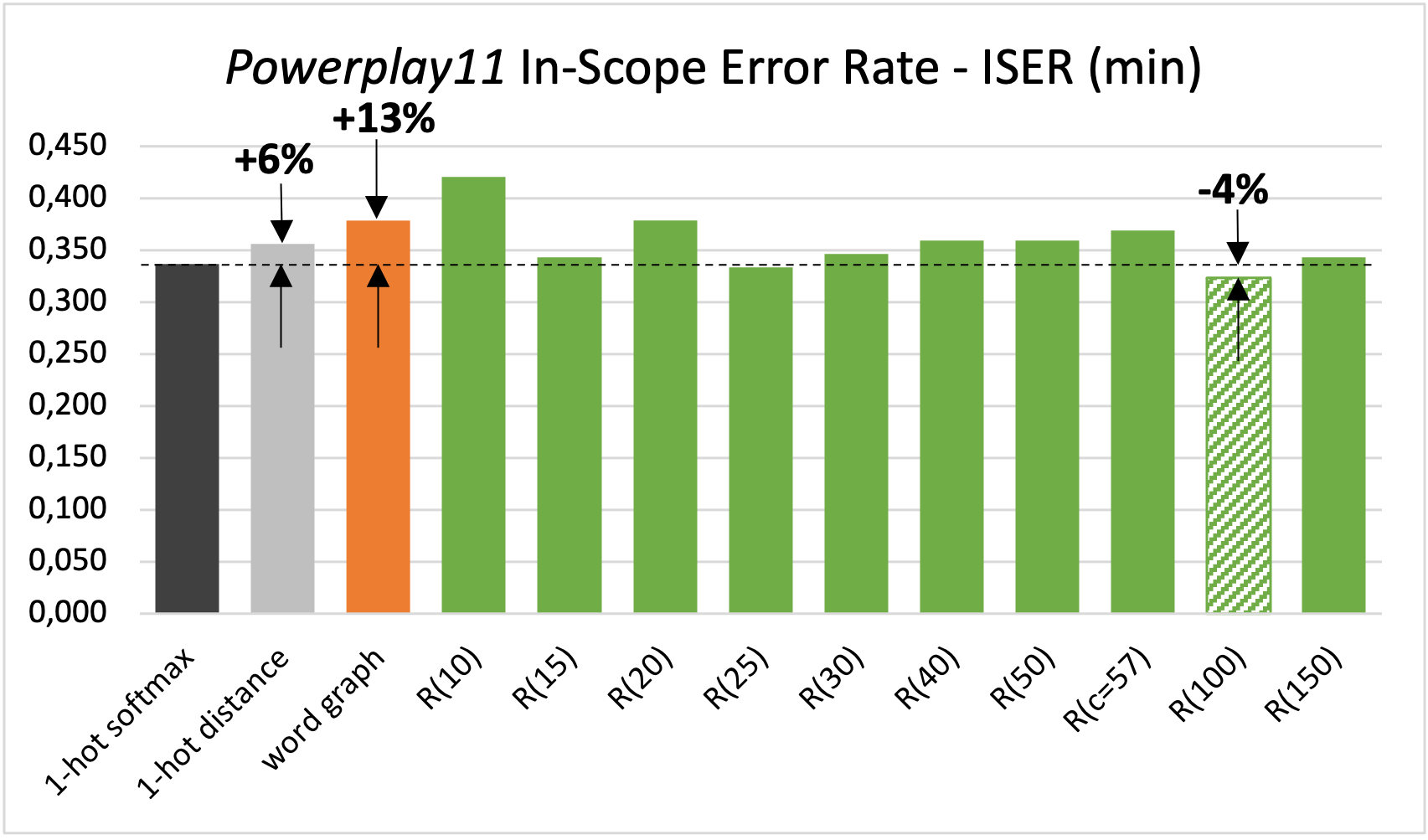} 
    \caption{Charts of FAR (min) and ISER (min) results on each of the HINT3 datasets. The difference percentage from \textit{1-hot softmax} is shown to \textit{1-hot distance}, to \textit{word graph}, and to the best $R(N)$ value, highlighted with diagonal stripes. }
    \label{fig:graph_HINT3}
\end{figure*}


\subsubsection{HINT3 Datasets}
The three HINT3 intent recognition problems~\cite{arora-etal-2020-hint3} consist of datasets containing actual user queries and real OOS examples. Each dataset is related to a single and unique domain, and the datasets possess real-world difficulties such as small and unbalanced training sets and a small number of intents. For the experiments, we considered only the \emph{full}  version of each of the three HINT3 datasets:
\begin{description}[nosep]
    \item{\textit{\textbf{SOFTMattress:}}} 21 intents, 328 training samples, and 231 IS and 166 OOS test samples;
    \item{\textit{\textbf{Curekart:}}} 28 intents, 600 training samples, and 452 IS and 539 OOS test samples;
    \item{\textit{\textbf{Powerplay11:}}} 59 intents, 471 training samples, and 275 IS and 708 OOS test samples.
\end{description}




\begin{table}[t]
    \centering
    \includegraphics[trim={0cm 2cm 0 0}, width=8.25cm]{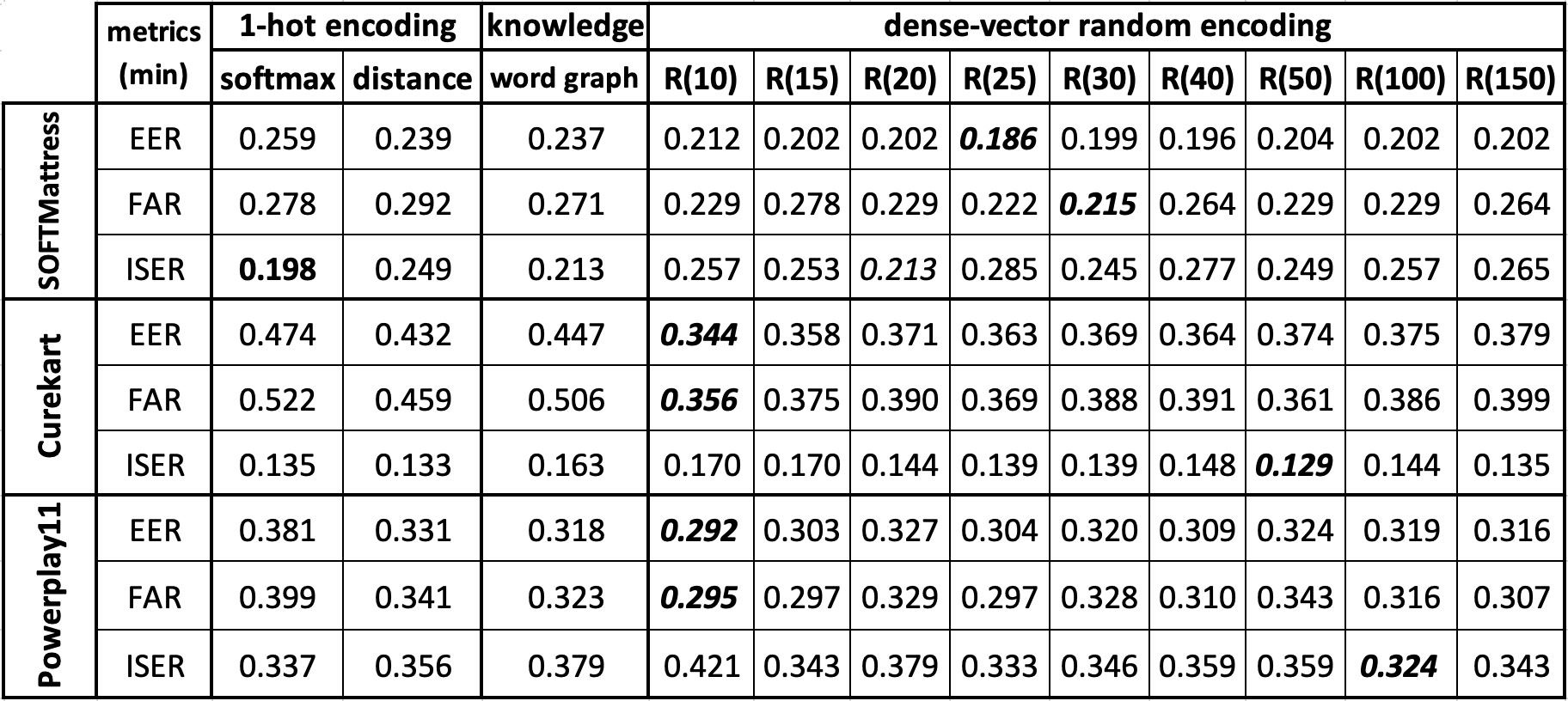}
    \caption{Summary of the EER, FAR, and ISER results on the HINT3  datasets showing the minimum (min) values. For each metric, the best result is in \textbf{bold} typeface; and, among the dense-vector encodings $R(N)$, in \textit{italic}. Full results available in Appendix~\ref{app:detailed_experiment_results}.}
    \label{tab:HINT3_results_summary}
\end{table}

The algorithms were configured as follows. In all methods, the neural network architecture had an input size of~512, the dimension of the USE vectors; one hidden layer with 768~neurons and dropout rate of~0.1; parameters trained with the \textit{Adam optimizer}; and minimizing categorical cross-entropy. For the \textit{word graph} method, we created the graph by finding the common words of the training examples as in~\cite{cavalin2020improving}. We then used \textit{DeepWalk} with the class embedding size set to~200 and walk sizes of~20. We used a two-layer neural network with 800 hidden neurons trained  for 50~epochs with the Adam optimizer.

We evaluated 10~different implementations of the $R(N)$ algorithm, with nine $N$ values ranging from 10 to 150, and one $N$ equal to the number of classes $c$. For each $R(N)$, we trained 500~different randomly generated class encodings to better explore the space of class encodings. Those systems were compared to one training of the \textit{1-hot softmax}, \textit{1-hot distance}, and \textit{word graph} algorithms. We are aware that differences in results can occur since the neural networks weights are randomly initialized but we found such differences negligible. Moreover, if we had to perform different training runs for the other three algorithms we would also need to perform different training runs for each of the 500~random class encodings.

Table~\ref{tab:HINT3_results_summary} shows a summary of the main results of the experiments on the HINT3 datasets, considering the minimum value of $R(N)$ in each metrics. More detailed results, including the average and standard deviation for the 500~random samplings, are available from table~\ref{tab:HINT3_results_full} in Appendix~\ref{app:detailed_experiment_results}. The best result for each metric is marked in bold typeface and the best result for the $R(N)$ algorithms in italic. 

Regarding the minimum EER values reached by the $R(N)$ methods, the results in table~\ref{tab:HINT3_results_summary} show that random sampling is promising towards finding good dense-vector encodings. For SOFTMattress, Curekart, and Powerplay11, the best overall EER values were 0.186, 0.344, and 0.292, respectively from $R(25)$, $R(10)$, and $R(10)$. Those correspond to improvements in EER, compared to \textit{1-hot softmax}, of -28\%, -27\%, and -23\%, respectively. 


The improvements in EER seemed to be chiefly due to improvements in the in the FAR metric of OOS detection. Table~\ref{tab:HINT3_results_summary} shows that gains in FAR happened in the three datasets for every $N$-dimension of $R(N)$, and in  $R(30)$, $R(10)$, and $R(10)$ we obtained gains in FAR of -23\%, -32\%, and -26\% to \textit{1-hot softmax}, respectively. This is highlighted by the top part of fig.~\ref{fig:graph_HINT3} which clearly shows that the FAR metrics of the best $R(N)$ encodings were not only significantly better than \textit{1-hot softmax}, but also beat easily \textit{1-hot distance} and \textit{word graph} in the three datasets. 


The same did not repeat for ISER, as seen in~table~\ref{tab:HINT3_results_summary}. When considering only the accuracy of classifying IS input in the intent classes,  dense-vector encodings were able to perform slightly better in Curekart and Powerplay11, with improvements of -5\% and -4\% but not in SOFTMattress. In the lower part of fig.~\ref{fig:graph_HINT3} we can clearly see the difficulty for almost all algorithms to be better than \textit{1-hot softmax}.



\begin{table}[t]
    \centering
    \includegraphics[trim={0cm 1cm 0 0}, width=8.25cm]{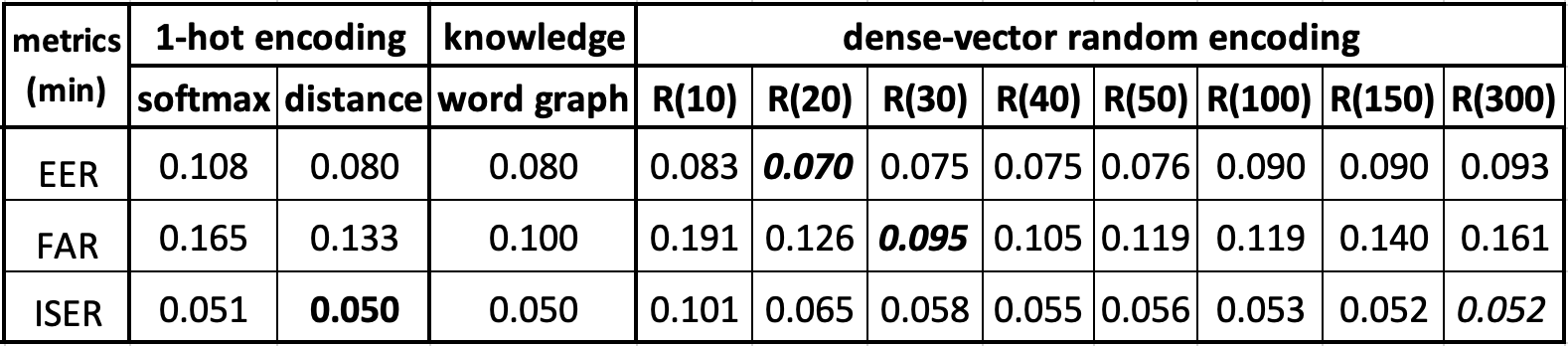} 
    \caption{Summary of the EER, FAR, and ISER results on the Larson  dataset showing the minimum (min) values. For each metric, the best result is in \textbf{bold} typeface; and, among the dense-vector encodings $R(N)$, in \textit{italic}. Full results available in Appendix~\ref{app:detailed_experiment_results}.}
    \label{tab:larson_results_summary}
\end{table}

\begin{figure}[t]
    \centering
    \includegraphics[width=5.5cm]{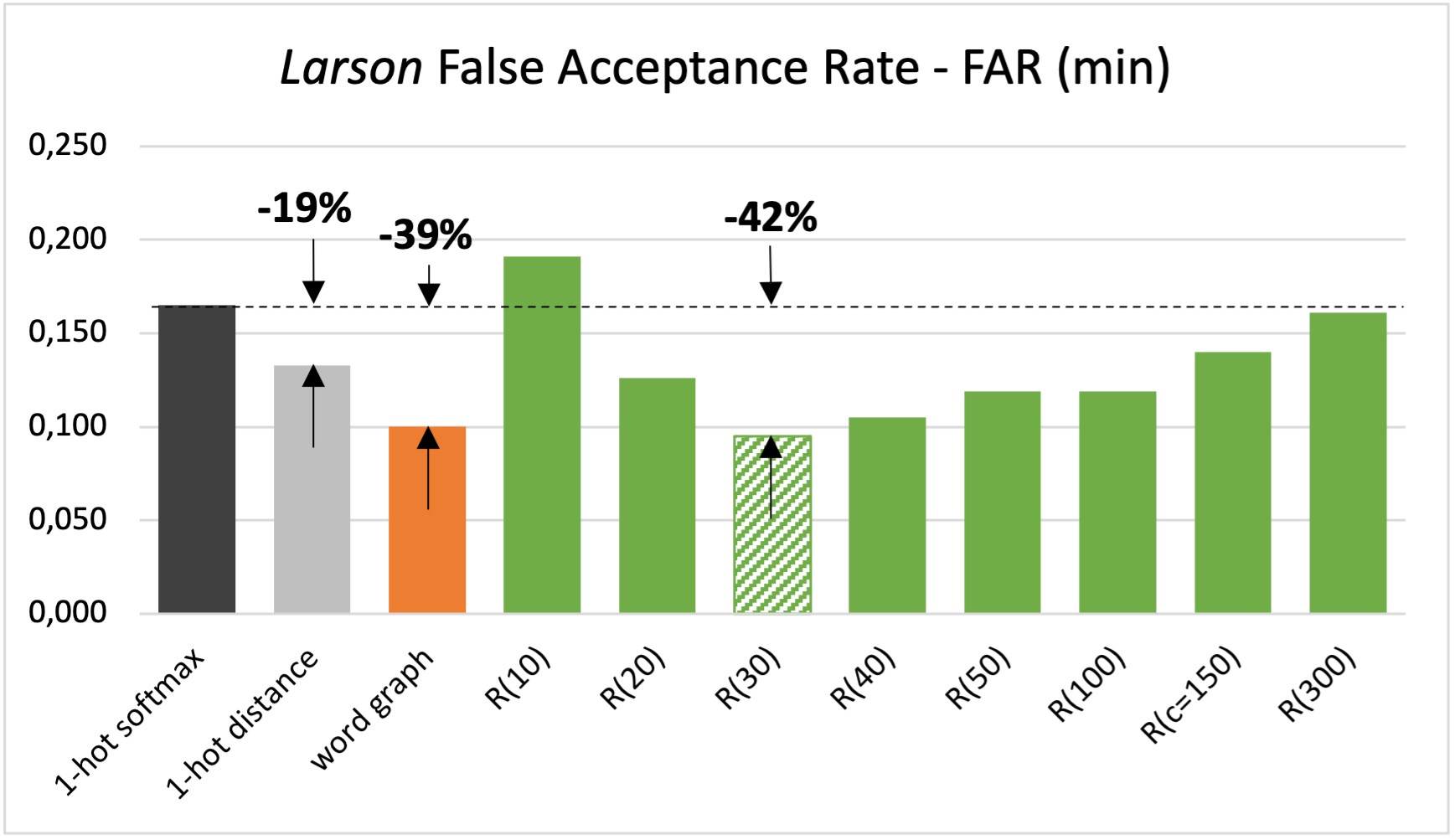} \\
    \includegraphics[width=5.5cm]{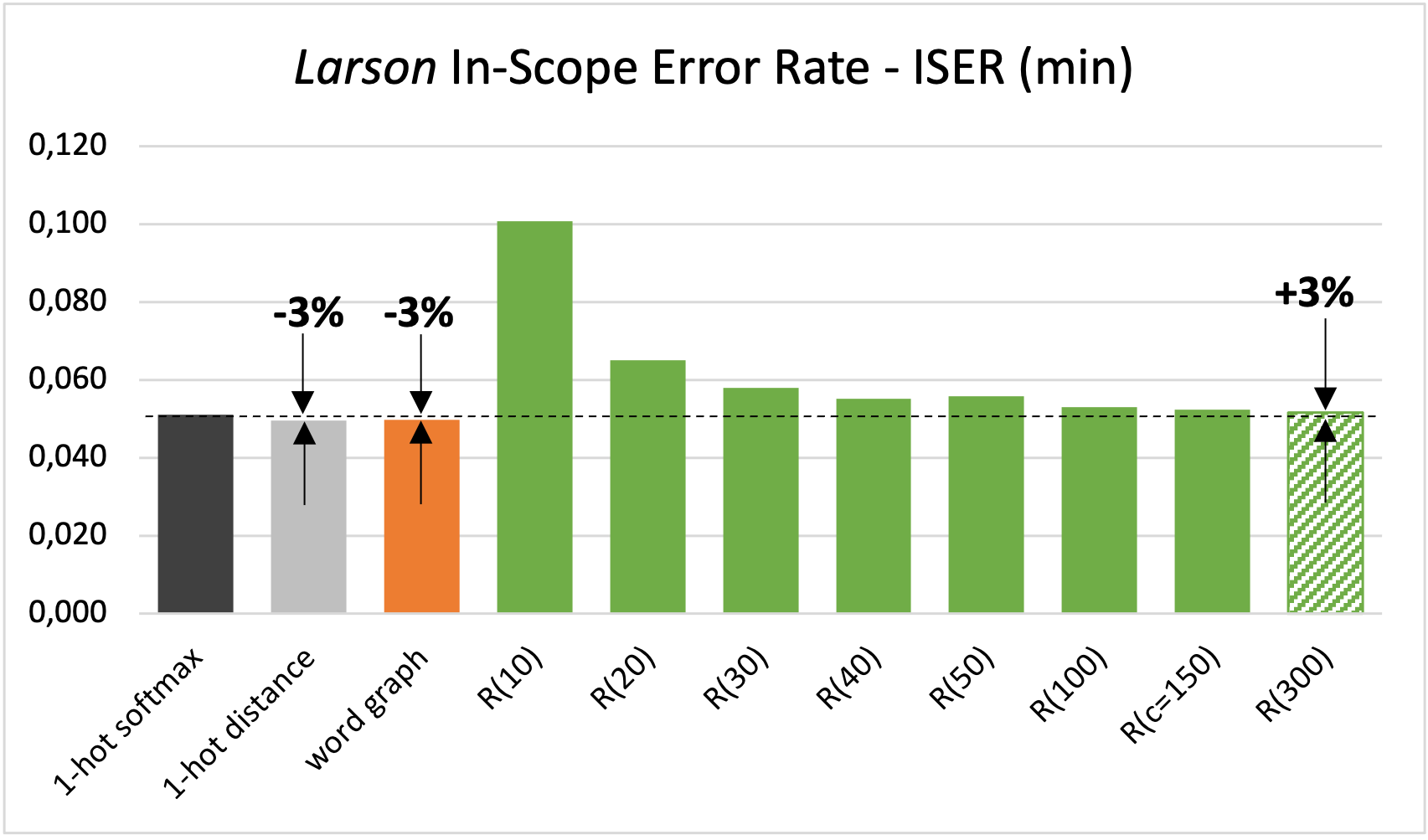}
    \caption{Charts of FAR (min) and ISER (min) results on the Larson dataset. The difference percentage from \textit{1-hot softmax} is shown to \textit{1-hot distance}, to \textit{word graph}, and to the best $R(N)$ value, highlighted with diagonal stripes.}
    \label{fig:graph_larson}
\end{figure}

\subsubsection{Larson Dataset}
We also explored the publicly-available~\textit{Larson} dataset~\cite{larson-etal-2019-evaluation}, a set specifically designed for the evaluation of the OOS detection. Unlike the HINT3 datasets, it contains a much larger number of samples with a total of 18,000 training samples and 5,500 test samples (4,500 IS and 1,000 OOS), and the number of examples per class is balanced. 
Moreover, the number of classes in this dataset is quite larger than in the HINT3 with 150 classes, 
and comprises five different domains unlike the single-domain HINT3 datasets. Also, the error rates reported in Larson are considerably lower, thus it is more challenging to achieve improvements to \textit{1-hot softmax}. Configuration-wise, we used a slightly modified set of values for $N$, starting in 10 and ending in 300, due to the larger number of classes. 

Table~\ref{tab:larson_results_summary} shows a summary of the main results of the experiments on the Larson dataset, considering the minimum value of $R(N)$ in each metrics. More detailed results are available from table~\ref{tab:larson_results_full} in Appendix~\ref{app:detailed_experiment_results}. 
In terms of EER, $R(20)$ was the best encoding, improving \textit{1-hot softmax} by -35\% and \textit{1-hot distance} and \textit{word graph} by -13\%. The improvement in FAR was even more impressive, from 0.165 of \textit{1-hot softmax} to 0.095 in $R(30)$, a massive gain of -42\%, although it was only slightly better than \textit{word graph} as seen in fig.~\ref{fig:graph_larson}. The ISER metrics was 3\% worse and for small dimensions it was twice as bad.

\section{Towards Optimal Class Encodings}
The results of our experiments indicate that there exist dense-vector encodings able to achieve better OOS detection results than  one-hot encodings and in some cases, than knowledge-based class encodings. This section proposes an algorithm to search for those ``optimal'' dense-vector encodings systematically.

\subsection{The Class Encoding Search (CES) Algorithm}

\begin{figure}[t]
    \centering
    \includegraphics[width=7cm]{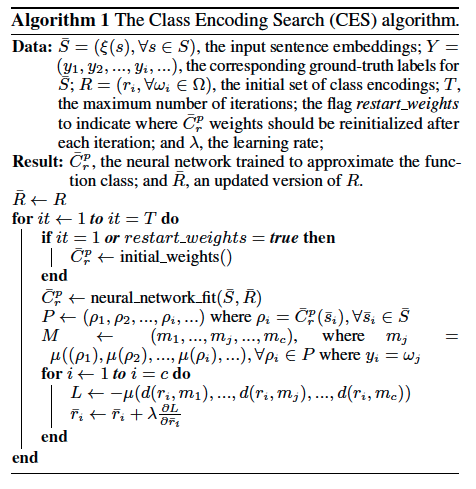}
    \caption{The Class Encoding Search (CES) algorithm.}
    \label{alg:class_encoding_search}
\end{figure}

We propose a \textit{Generative Adversarial Network (GAN)}-inspired algorithm which improves class encodings after a finite number of iterations. The algorithm, called \textit{Class Encoding Algorithm (CES)} and detailed in Algorithm~\ref{alg:class_encoding_search}, takes as input a given initial class encoding computed, for instance, by a random sampling. The goal is to find an optimal encoding for $R(N=p)$, i.e. a instance of the $\bar{C}_{r}^{p}$ function, which is accurate at both OOS detection (FAR) and in-scope sample classification (ISER). 

Considering that $W_{\bar{C}}$ represents the set of weights used by $\bar{C}_{r}^{p}$ and that during training there is no OOS example available, our GAN-inspired algorithm  consists of continuously updating $W_{\bar{C}}$ and $R(p)$ to achieve two distinct but complementary objectives: to create a function class $\bar{C}_{r}^{p}$ which projects the sentences in $S$ onto compact spheres, while considering class encodings where each point is the farthest way from the other ones. That is achieved by computing $W_{\bar{C}}$ by minimizing the mean squared error and by updating $R$ by maximizing the distance of each point to the others. A similar dual-objective optimization has shown promising results in \cite{zeng-etal-2021-modeling}. For this we use a $\mu(X)$ function which computes the mean 1D vector from a set of 1D vectors with the same dimension represented by $X$.

\begin{figure}[t]
    \centering
    \includegraphics[width=8cm]{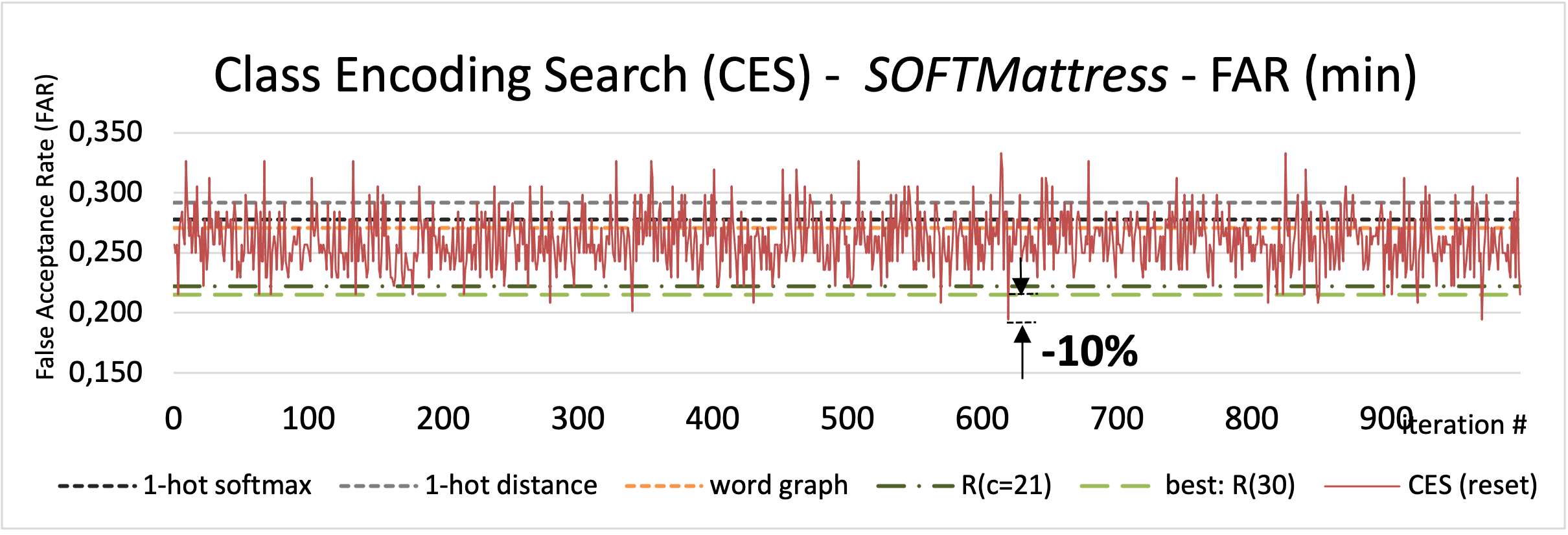}
    \\ \vspace{1mm} 
    \includegraphics[width=8cm]{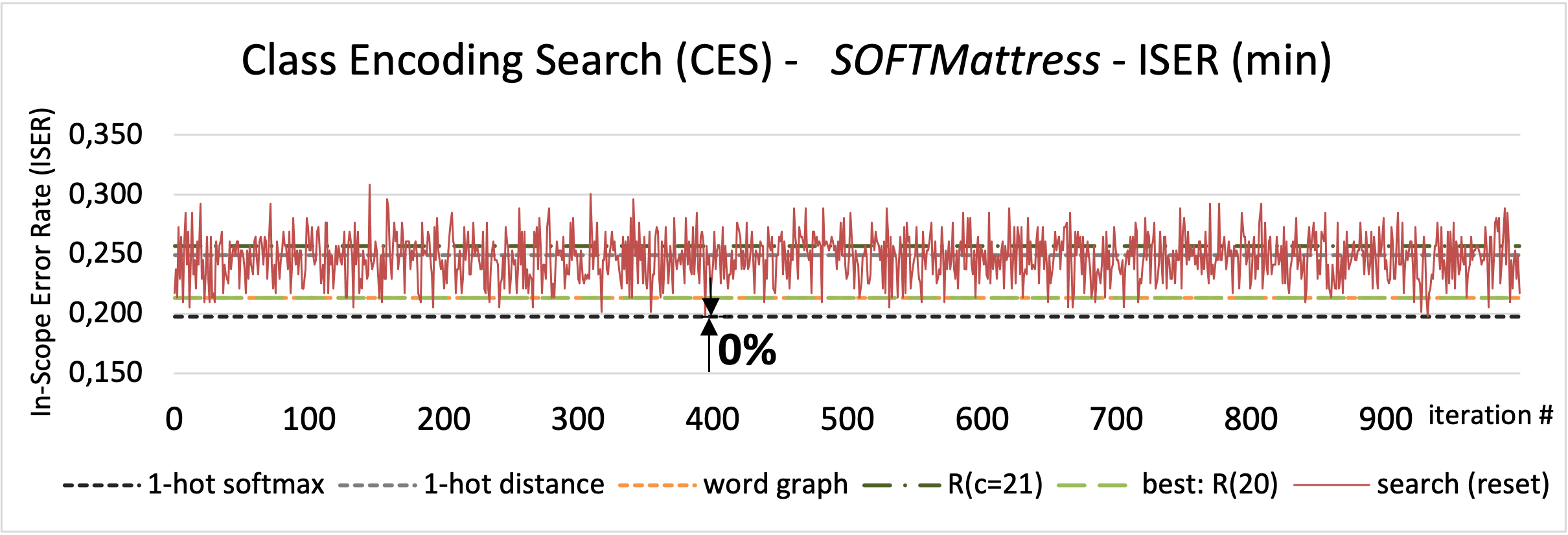}
    \caption{FAR and ISER results of the Class Embedding Search algorithm on the SOFTMattress dataset, during 1,000 iterations, $restart{\_}weights = true$.}
    \label{fig:ces_softmattress1k}
\end{figure}

\subsection{Initial Experiments with SOFTMattress}
We applied the CES Algorithm on the SOFTMattress dataset, considering $R(N = c)$ as the base class encoding approach. Class encodings were updated to 1,000 iterations, and we experimented the $restart{\_}weights$ argument set to both $true$ and $false$. In the first case,  a new $W_{\bar{C}}$ set of weights for $\bar{C}_{r}^{p}$ is trained from scratch at the beginning of each iteration. In the second one, $W_{\bar{C}}$ is only initialized in the first iteration, and the weights are only updated in the subsequent iterations. In our experiments, 
the learning rate $\lambda$ was set to 0.0001 after some preliminary experiments.


The results for $restart\_weights = true$ are depicted in fig.~\ref{fig:ces_softmattress1k}. For the FAR metric we observed a considerable impact, with the algorithm finding a class embedding at iteration~$619$ where the FAR value dropped from the previous best result of 0.215 with $R(30)$ to 0.194, a reduction of about 10\%. However, we did not see a pattern of convergence to a minimum which would be highly desirable. For the ISER metric, the algorithm found at iteration~$394$ an ISER value of 0.198, as good as the best ISER value obtained by the method using the \textit{1-hot softmax} encoding. Notice that this was a reduction of about 7\% over the best previous $p-$dimensional spatial encoding, $R(20)$.
We see both results as very promising about the ability of finding good class encodings by the CES algorithm but acknowledge that further development is needed to assure that the results converge to the minimum values.

\section{Conclusions}
In this work we presented an investigation of dense-vector class embeddings in the context of intent classification with OOS detection. We first showed that dense-vector encodings allow much more complex OOS spaces in terms of the number of different topologies they can represent. We then performed experiments in four datasets for OOS detection, where we found that we could find good dense-vector representations by random sampling. Using the FAR metric of OOS detection, we saw improvements between 20\% and 40\% over typical one-hot softmax encodings of outputs and obtaied better results than the knowledge-based SOTA method, but got no significant gains in-scope error rate (ISER). We then presented a GAN-inspired search algorithm and showed, in preliminary experiments, that it could find dense-vector encodings which were better than the ones we had found by random sampling.

We believe both the theoretical exploration and the experimental results support the importance of using dense-vector encoding of intent classes in classification with OOS detection, that is, in open world scenarios of intent classification. At the same time, these results question the need of knowledge-based class encoding as argued in~\cite{cavalin2020improving,pinhanez-etal-2021-using}. Of course, if such knowledge is available, it should be used, and this work has shown only indications that it can be significantly improved. But when such knowledge is not immediately present or minable, our proposed algorithm can search simultaneously the space of encodings and neural networks to find a good encoding.


More broadly, our results suggest that ML systems built  using one-hot encoding may be improved, perhaps dramatically, in the case of open world scenarios by switching to dense-vector encodings of the output layer, with or without domain knowledge. Investigating further this possibility is imperative, especially in domains other than intent classification, as well as understanding its underlying mechanisms.

\pagebreak
\bibliography{bibliography}
\bibliographystyle{icml2022}

\vfill\null

\pagebreak 

\vfill\null

\pagebreak

\appendix

\section{Class Encoding Topologies in Higher Dimensions}\label{app:class_encoding_topologies_higher_dimensions}

In section~\ref{sec:class_encodings_topologies} we explored the different topologies in 2D afforded by the class encodings described  section~\ref{sec:class_encoding_oos}. In this section we show how those results extend to greater dimensions, that is, $c>2$. As before, we do not consider here special, limit cases where the intersection of two components degenerates to a single point or to a tangent line, and only the cases where OOS detection is needed, that is, $\theta>0$. 

\textbf{Topologies of $\mathbf{\bar{C}_{max}}$:} let us consider the $\bar{C}_{max}$ function, with $c>2$ classes $\omega_1, \omega_2, ...\omega_c$ being recognized, or the symbol $o$ of a OOS input being produced. It is easy to see that the OOS class occupies a $c$-dimensinal hypercube and each class $\omega_i$ a $c$-dimensional hyper-trapezium. Moreover, each pair of components corresponding to the $\omega_i$ and $\omega_j$ classes are in contact through the diagonal of the plane defined by the axis of the coordinates $z_i$ and $z_j$. Similarly, each triad $\omega_i$, $\omega_j$, and $\omega_k$ classes share the diagonal of the hyper-cube defined by the coordinates $z_i$, $z_j$, and $z_k$. And the same is valid for every subset $\{\omega_{i_1}, \omega_{i_2}, ...\omega_{i_t}\}$ of $\Omega$, $2 \leq t \leq c$. Therefore, all pair of classes are connected to each other, and so all triads of classes, and so on, in a $c$-dimensional complete hypergraph, and thus, like in the 2D case, \textbf{the $\bar{C}_{max}$ function defines only 1 topology for any number of classes $c \geq 2$}.

\textbf{Topologies of $\mathbf{\bar{C}_{softmax}}$:} let us consider the case of the $\bar{C}_{softmax}$ function. In the 2D case, there were two possible topologies, defined by $0 < \theta \leq 1/2$ and $1/2 < \theta \leq 1$. We show here that, for any $c$, the $\bar{C}_{softmax}$ defines exactly $c$ different topologies, corresponding to the intervals $0 < \theta \leq 1/c$, $1/c < \theta \leq 1/(c-1)$, ... , $1/2 < \theta \leq 1$. We showed in section~\ref{sec:class_encodings_topologies} that in the case where $c=2$, there are exactly two topologies, one where all classes are connected when $0 < \theta \leq 1/2$ and one where all classes are disconnected when $1/2 < \theta \leq 1$. Using induction, let us assume that it is true that the number of topologies in the case of $c>2$ classes is exactly $c$, defined by the intervals $0 < \theta \leq 1/c$, $1/c < \theta \leq 1/(c-1)$, ... , $1/2 < \theta \leq 1$. Let us consider the case of $c+1$.
First, let us observe that the $1/2 < \theta \leq 1$ in the $c$ case generates a topology where all $c$ classes are disconnected. It is easy to see that this holds in the $c+1$-dimensional space, and so it goes to all intervals until $1/c < \theta \leq 1/(c-1)$. While the interval $0 < \theta \leq 1/c$ in the $c$ case generated a topology where all classes are fully connected, that is not true in the case in $c+1$. It is necessary to split the interval $0 < \theta \leq 1/c$ into $0 < \theta \leq 1/(c+1)$ and $1/(c+1) < \theta \leq 1/c$. In the interval $1/(c+1) < \theta \leq 1/c$ all subsets of $c$ components or less are hyperconnected, but only when $0 < \theta \leq 1/(c+1)$, all $c+1$ classes have points in common, corresponding to the additional topology. Since each of the $c+1$ intervals is associated to a different topology, there are $c+1$ topologies, completing the induction. Therefore \textbf{the $\bar{C}_{softmax}$ function defines $c$ different topologies when the number of classes is $c$}.

\textbf{Topologies of $\mathbf{\bar{C}_{d}}$:} The case of the one-hot softmax is analogous to the one-hot encoding using the Euclidian distance. While in the former the intervals are defined by $1/c$, in the latter they are defined by $1/\sqrt{c}$, or $\sqrt{c}/c$. The same reasoning used in the case of $\bar{C}_{softmax}$ can show that $\bar{C}_{d}$ defines exactly $c$ different topologies, corresponding to the intervals $0 < \theta \leq 1/\sqrt{c}$, $1/\sqrt{c} < \theta \leq 1/\sqrt{c-1}$, ... , $1/\sqrt{2} < \theta \leq 1$. Therefore \textbf{the $\bar{C}_{d}$ function defines $c$ different topologies when the number of classes is $c$}.

\textbf{Topologies of $\mathbf{\bar{C}_{r}}$:} In section~\ref{sec:class_encodings_topologies} we showed that the $\bar{C}_{r}$ affords 12~different topologies when $c=2$. For the sake of the argument of this paper, that is, that dense-vector class encodings enable richer representations of the space of classes, it is not necessary to provide an exact estimation of the topologies defined by $\bar{C}_{r}$ in the case of $c$ classes. We here simply shows that the number of different topologies in this case grows quadratically with the number of classes by demonstrating that the number of different topologies $M_c$ is equal or greater than the sum $\sum_{m=1}^c m = m(m+1)/2$. We have shown that this is true in the case of $c=2$, which defines 12~topologies, clearly greater than $\sum_{m=1}^2 m = 1+2 = 3$. Using again induction, let us assume that if the inequality holds for $c$, that is, if $\bar{C}_{r}$ defines at least $M_c \geq \sum_{m=1}^c m$ different topologies when there are $c$ classes, then it is also true for $c+1$. To do so, first observe that all $M$ topologies can be extended to the $c+1$ dimension by considering a value of $-1$ for the $c+1$ coordinate of each $r_i$ dense vector, $1 \leq i \leq c$. If we define the $r_{c+1}$ to be the one-hot vector of $c+1$ dimension, it is easy to see that the $c+1$ component is disjoint with all other components, since $\theta <1$. Therefore there are at least $M_c$ different topologies defined by the $c+1$-dimensional $\bar{C}_{r}$, all of them with the $\omega_{c+1}$ component disjoint from the other components. Therefore, we just need to show that there are $c+1$ other topologies to complete this demonstration. We first show that it is possible to define $c$ different vectors $r_{c+1}$, denoted here as $r_{c+1}^i$, in such a way that its associated component is connected to only one of the other vectors. In fact, given $r_i=(r_{i1}, r_{i2}, ... r_{ic})$, $r_{ij} \in [-1,1]^c$, we can extend all $r_i$ to the $c+1$ dimension by setting the $c+1$ coordinate to $-\theta_i$, $\theta_i < 1$, and for each $r_i$, consider a vector $r_{c+1}^i$ such as $r_{c+1}^i=(r_{i1}, r_{i2}, ... r_{ic},\theta_i)$. It is easy to see that each of the $c$ spaces combining the original $r_1,r_2,...r_c$ dense-vectors to each of the $r_{c+1}^i$ dense-vector has at least one new topology where the component $\omega_{c+1}$ is connected to one component $\omega_i, i \leq c$. We can always finding a $\theta_i$ which is large enough so the component $\omega_{c+1}$ is connected only to the component $\omega_i$, there warranting that all the $c$ encodings constructed this way have different topologies among themselves and also different from any the previous $M_c$ topologies (where the component of $\omega_{c+1}$ was always not connected to any other component). We need only to demonstrate that we can create one more topology, different from all of the previous ones, to finish. To do so, let us extend all $r_i$ to the $c+1$ dimension by setting the $c+1$ coordinate to $0$. If we define $r_{c+1}$ at the center of the space, that is, with all coordinates as $0$, it is easy to notice that there is a $0 < \theta < 1$ which the hyper-dimension ``ball'' of radius $\theta$ centered in $r_{c+1}$ intersects with at least two other ``balls'' of radius $\theta$ defined by $r_i$ and $r_j$, $1\leq i,j \leq c$. Therefore the set of dense vectors defined this way is guaranteed to define a topology where the component $\omega_{c+1}$ is connected to at least two other components, and thus is different from the ones we constructed before. Thus, $M_{c+1} \geq M_c + c + 1$, but since $\sum_{i=1}^{c+1} = \sum_{i=1}^{c} + (c+1)$, we obtain $M_{c+1} \geq \sum_{i=1}^{c+1}$. As observed before, $\sum_{i=1}^{c+1} > (c+1)^2$, yielding $M_{c+1} \geq (c+1)^2$, and completing the demonstration. Therefore \textbf{the $\bar{C}_{r}$ function defines more than $c^2$ different topologies when the number of classes is $c$}.

This completes the demonstration that, for any number of classes $c \geq 2$, the one-hot encoding function $\bar{C}_{max}$ defines only one topology; the $\bar{C}_{softmax}$ and $\bar{C}_d$ define exactly $c$ different topologies; and for dense-vector encoding methods such as $\bar{C}_r$, the number of different topologies increases at least quadratically with $c$, as stated in section~\ref{sec:class_encodings_topologies}.

\section{Detailed Experiment Results}
\label{app:detailed_experiment_results}

\begin{table*}[t!]
    \centering
    \includegraphics[width=16cm]{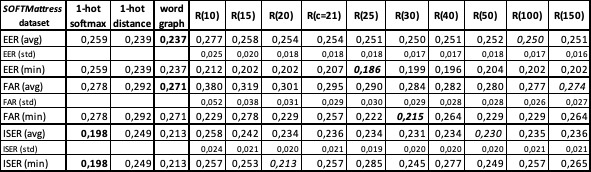} \\ \vspace{2mm}
    \includegraphics[width=16cm]{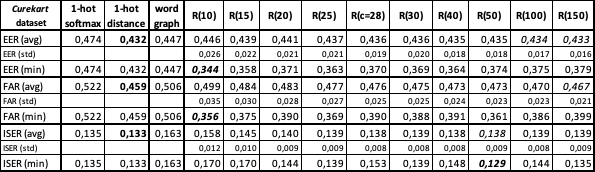} \\ \vspace{2mm}
    \includegraphics[width=16cm]{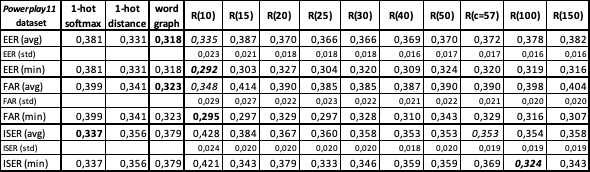} 
    \caption{Full experimental results of the EER, FAR, and ISER metrics on the HINT3  datasets showing the average (avg), its standard deviation (std), andminimum (min) values. For each metric, the best result is in \textbf{bold} typeface; and, among the dense-vector encodings $R(N)$, in \textit{italic} typeface. These resuls are summarized in table ~\ref{tab:HINT3_results_summary}}
    \label{tab:HINT3_results_full}
\end{table*}

This section presents more detailed results of the experiment described in section~\ref{sec:experiments_results}.

Table~\ref{tab:HINT3_results_full} expands table~\ref{tab:HINT3_results_summary} showing all the results associated to the experiments with the HINT3 datasets, including the average results for each metric and its associated standard deviation. We can see that, most often, the dense-vector encoding with minimum value of a metric produced a considerably smaller value for that metric. In other words, finding a dense-vector encoding which minimizes the metrics is not a trivial task.

\begin{table*}[t!]
    \centering
    \includegraphics[width=16cm]{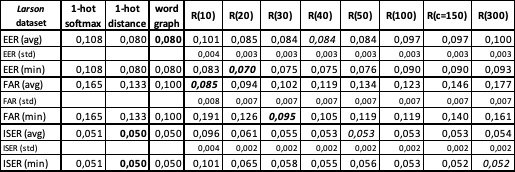} 
    \caption{Full experimental results of the EER, FAR, and ISER metrics on the Larson dataset showing the average (avg), its standard deviation (std), andminimum (min) values. For each metric, the best result is in \textbf{bold} typeface; and, among the dense-vector encodings $R(N)$, in \textit{italic} typeface. These resuls are summarized in table ~\ref{tab:larson_results_summary}.}
    \label{tab:larson_results_full}
\end{table*}

Table~\ref{tab:larson_results_full} expands table~\ref{tab:larson_results_summary} showing all the results associated to the experiments with the Larson datasets, including the average results for each metric and its associated standard deviation. Similar to the HINT3 datasets, most often, the dense-vector encoding with minimum value of a metric produced a considerably smaller value for that metric. In other words, finding a dense-vector encoding which minimizes the metrics is not a trivial task.

\vfill\null 

\end{document}